\documentclass{article}

\usepackage{times}
\usepackage[dvipdfmx]{graphicx}

\usepackage{natbib}
\usepackage[caption=false,font=footnotesize]{subfig}

\usepackage{algorithm}
\usepackage{algorithmic}
\def\vector#1{\mbox{\boldmath $#1$}}

\usepackage{listings}
\usepackage{amsmath}
\usepackage{amssymb}
\usepackage{amsthm}
\usepackage{mathtools}
\usepackage{comment}

\theoremstyle{plain}

\DeclareMathOperator*{\argmax}{arg\,max}

\newcounter{long}
\usepackage{hyperref}

\usepackage[accepted]{icml2017}

\icmltitlerunning{Learning Discrete Representations via Information Maximizing Self-Augmented Training}

\begin{document} 

\twocolumn[
\icmltitle{Learning Discrete Representations via Information Maximizing \\ Self-Augmented Training}



\icmlsetsymbol{equal}{*}

\begin{icmlauthorlist}
\icmlauthor{Weihua Hu}{tokyo,riken}
\icmlauthor{Takeru Miyato}{pfn,atr}
\icmlauthor{Seiya Tokui}{pfn,tokyo}
\icmlauthor{Eiichi Matsumoto}{pfn,tokyo}
\icmlauthor{Masashi Sugiyama}{riken,tokyo}
\end{icmlauthorlist}

\icmlaffiliation{tokyo}{University of Tokyo, Japan}
\icmlaffiliation{pfn}{Preferred Networks, Inc., Japan}
\icmlaffiliation{atr}{ATR Cognitive Mechanism Laboratories, Japan}
\icmlaffiliation{riken}{RIKEN AIP, Japan}

\icmlcorrespondingauthor{Weihua Hu}{hu@ms.k.u-tokyo.ac.jp}
\icmlcorrespondingauthor{Takeru Miyato}{takeru.miyato@gmail.com}

\icmlkeywords{boring formatting information, machine learning, ICML}

\vskip 0.3in
]



\printAffiliationsAndNotice{} 

\begin{abstract} 
Learning discrete representations of data is a central machine learning task because of the compactness of the representations and ease of interpretation. 
The task includes clustering and hash learning as special cases. 
Deep neural networks are promising to be used because they can model the non-linearity of data and scale to large datasets.
However, their model complexity is huge, and therefore, we need to carefully regularize the networks in order to learn useful representations that exhibit intended invariance for applications of interest.
To this end, we propose a method called Information Maximizing Self-Augmented Training (IMSAT). 
In IMSAT, we use data augmentation to impose the invariance on discrete representations.
More specifically, we encourage the predicted representations of augmented data points to be close to those of the original data points in an end-to-end fashion.
At the same time, we maximize the information-theoretic dependency between data and their predicted discrete representations.
Extensive experiments on benchmark datasets show that IMSAT produces state-of-the-art results for both clustering and unsupervised hash learning.



\end{abstract} 

\section{Introduction}
\begin{figure}[ht]\label{fig:image}
\begin{center}
\centerline{\includegraphics[width=7cm, clip]{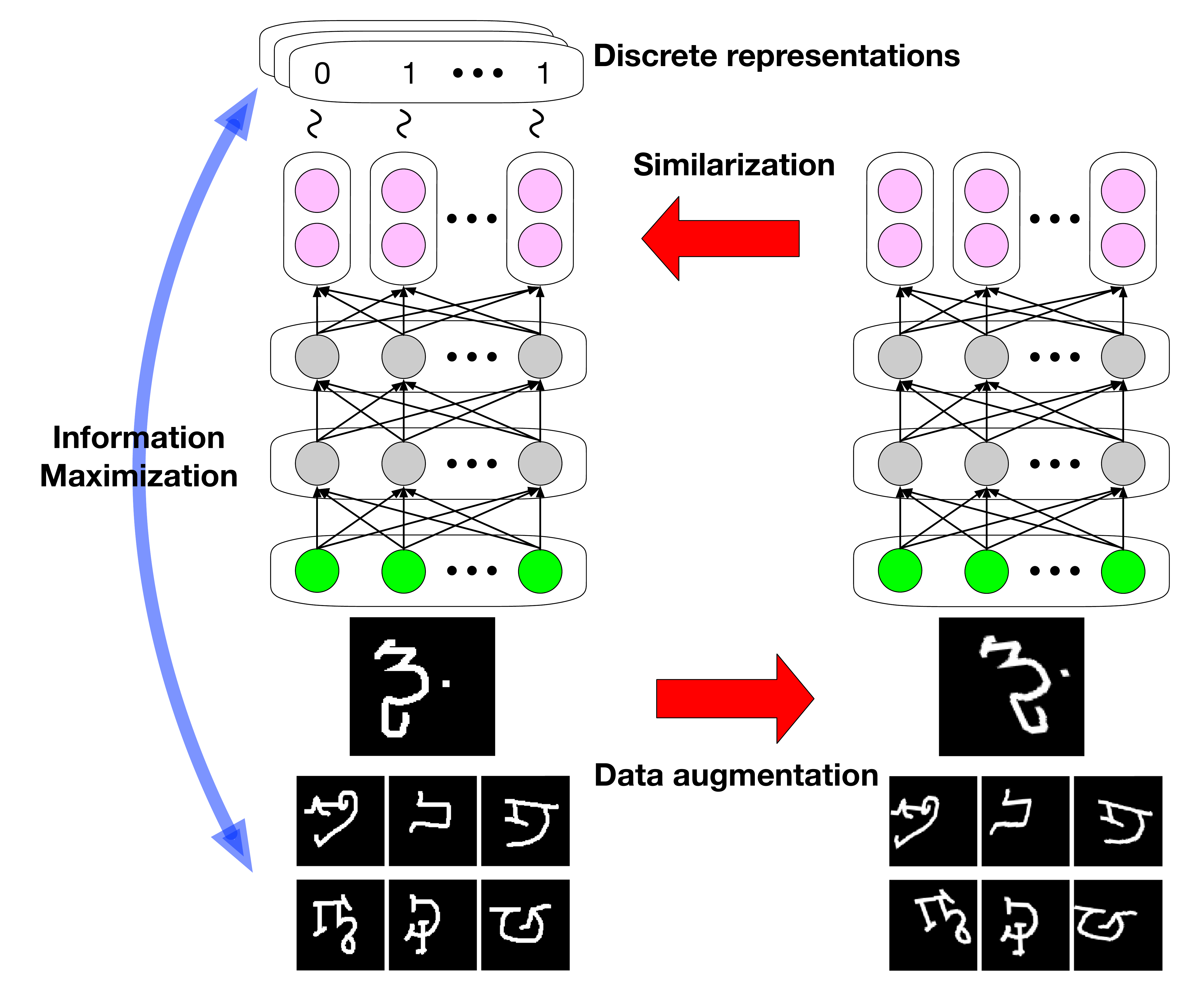}}
\caption{Basic idea of our proposed method for unsupervised discrete representation learning. We encourage the prediction of a neural network to remain unchanged under data augmentation (Red arrows), while maximizing the information-theoretic dependency between data and their representations (Blue arrow).}
\end{center}
\vspace{-1cm}
\end{figure}

The task of unsupervised discrete representation learning is to obtain a function that maps \emph{similar} (resp.~\emph{dissimilar}) data into similar (resp.~dissimilar) discrete representations, where the \emph{similarity} of data is defined according to applications of interest.
It is a central machine learning task because of the compactness of the representations and ease of interpretation.  
The task includes two important machine learning tasks as special cases: clustering and unsupervised hash learning. 
Clustering is widely applied to data-driven application domains \citep{berkhin2006survey}, while hash learning is popular for an approximate nearest neighbor search for large scale information retrieval \citep{wang2016learning}. 

Deep neural networks are promising to be used thanks to their scalability and flexibility of representing complicated, non-linear decision boundaries.
However, their model complexity is huge, and therefore, regularization of the networks is crucial to learn meaningful representations of data.
Particularly, in \emph{unsupervised} representation learning, target representations are not provided and hence, are unconstrained.
Therefore, we need to carefully regularize the networks in order to learn useful representations that exhibit intended invariance for applications of interest (e.g., invariance to small perturbations or affine transformation).
Na\"{\i}ve regularization to use is a weight decay \citep{erin2015deep}.
Such regularization, however, encourages global smoothness of the function prediction; thus, may not necessarily impose the intended invariance on the predicted discrete representations.

Instead, in this paper, we use data augmentation to model the invariance of learned data representations.
More specifically, we map data points into their discrete representations by a deep neural network and regularize it by encouraging its prediction to be invariant to data augmentation.
The predicted discrete representations then exhibit the invariance specified by the augmentation.
Our proposed regularization method is illustrated as red arrows in Figure~\ref{fig:image}.
As depicted, we encourage the predicted representations of augmented data points to be close to those of the original data points in an end-to-end fashion. 
We term such regularization \emph{Self-Augmented Training (SAT)}.
SAT is inspired by the recent success in regularization of neural networks in semi-supervised learning \citep{bachman2014learning, miyato2015distributional, sajjadi2016regularization}.
SAT is flexible to impose various types of invariances on the representations predicted by neural networks.
For example, it is generally preferred for data representations to be locally invariant, i.e., remain unchanged under local perturbations on data points.
Using SAT, we can impose the local invariance on the representations by pushing the predictions of perturbed data points to be close to those of the original data points. 
For image data, it may also be preferred for data representations to be invariant under affine distortion, e.g., rotation, scaling and parallel movement.
We can similarly impose the invariance via SAT by using the affine distortion for the data augmentation.

We then combine the SAT with the Regularized Information Maximization (RIM) for clustering \citep{krause2010discriminative, heading1991unsupervised}, and arrive at our Information Maximizing Self-Augmented Training (IMSAT), an information-theoretic method for learning discrete representations using deep neural networks. 
We illustrate the basic idea of IMSAT in Figure~\ref{fig:image}. 
Following the RIM, we maximize information theoretic dependency between inputs and their mapped outputs, while regularizing the mapping function.
IMSAT differs from the original RIM in two ways.
First, IMSAT deals with a more general setting of learning discrete representations; thus, is also applicable to hash learning.
Second, it uses a deep neural network for the mapping function and regularizes it in an end-to-end fashion via SAT.
Learning with our method can be performed by stochastic gradient descent (SGD); thus, scales well to large datasets.

In summary, our contributions are: 1) an information-theoretic method for unsupervised discrete representation learning using deep neural networks with the end-to-end regularization, and 2) adaptations of the method to clustering and hash learning to achieve the state-of-the-art performance on several benchmark datasets.

The rest of the paper is organized as follows. 
Related work is summarized in Section \ref{sec:related}, while our method, IMSAT, is presented in Section \ref{sec:method}. 
Experiments are provided in Section \ref{sec:experiment} and conclusions are drawn in Section \ref{sec:conclusion}.

\section{Related Work} \label{sec:related}

Various methods have been proposed for clustering and hash learning.
The representative ones include $K$-means clustering and hashing \citep{he2013k}, Gaussian mixture model clustering, iterative quantization \citep{gong2013iterative}, and minimal-loss hashing \citep{norouzi2011minimal}. 
However, these methods can only model linear boundaries between different representations; thus, cannot fit to non-linear structures of data.
Kernel-based \citep{xu2004maximum, kulis2009learning} and spectral \citep{ng2001spectral, weiss2009spectral} methods can model the non-linearity of data, but they are difficult to scale to large datasets. 

Recently, clustering and hash learning using deep neural networks have attracted much attention. 
In clustering, \citet{xie2016unsupervised} proposed to use deep neural networks to simultaneously learn feature representations and cluster assignments, while \citet{dilokthanakul2016deep} and \citet{zheng2016variational} proposed to model the data generation process by using deep generative models with Gaussian mixture models as prior distributions. 

Regarding hashing learning, a number of studies have used deep neural networks for supervised hash learning and achieved state-of-the-art results on image and text retrievals \citep{xia2014supervised, lai2015simultaneous, zhang2015bit, xu2015convolutional, li2015feature}. 
Relatively few studies have focused on unsupervised hash learning using deep neural networks. 
The pioneering work is semantic hashing, which uses stacked RBM models to learn compact binary representations \citep{salakhutdinov2009semantic}. 
\citet{erin2015deep} recently proposed to use deep neural networks for the mapping function and achieved state-of-the-art results. 
These unsupervised methods, however, did not explicitly intended impose the invariance on the learned representations. 
Consequently, the predicted representations may not be useful for applications of interest.

In supervised and semi-supervised learning scenarios, data augmentation has been widely used to regularize neural networks. \citet{leen1995data} showed that applying data augmentation to a supervised learning problem is equivalent to adding a regularization to the original cost function. 
\citet{bachman2014learning, miyato2015distributional, sajjadi2016regularization} showed that such regularization can be adapted to semi-supervised learning settings to achieve state-of-the-art performance. 

In unsupervised representation learning scenarios, \citet{dosovitskiy2014discriminative} proposed to use data augmentation to model the invariance of learned representations.
Our IMSAT is different from \citet{dosovitskiy2014discriminative} in two important aspects: 1) IMSAT \emph{directly} imposes the invariance on the learned representations, while \citet{dosovitskiy2014discriminative} imposes invariance on surrogate classes, \emph{not directly} on the learned representations.~2) IMSAT focuses on learning \emph{discrete} representations that are directly usable for clustering and hash learning, while \citet{dosovitskiy2014discriminative} focused on learning \emph{continuous} representations that are then used for other tasks such as classification and clustering.
\ifnum\value{long}<1
Relation of our work to denoising and contractive auto-encoders \citep{vincent2008extracting, rifai2011contractive} is discussed in Appendix \ref{app:dae} of the supplementary material. 
\else
Relation of our work to denoising and contractive auto-encoders \citep{vincent2008extracting, rifai2011contractive} is discussed in Appendix \ref{app:dae}. 
\fi

\section{Method}\label{sec:method}
Let $\mathcal{X}$ and $\mathcal{Y}$ denote the domains of inputs and discrete representations, respectively. 
Given training samples, $\{x_1, x_2, \ldots, x_N \}$, the task of discrete representation learning is to obtain a function, $f: \mathcal{X} \to \mathcal{Y}$, that maps similar inputs into similar discrete representations. The similarity of data is defined according to applications of interest.

We organize Section \ref{sec:method} as follows.
In Section \ref{subsec:rim}, we review the RIM for clustering \citep{krause2010discriminative}.  
In Section \ref{subsec:imsat}, we present our proposed method, IMSAT, for discrete representation learning.
In Sections \ref{subsec:adapt_clustering} and \ref{subsec:adapt_hashing}, we adapt IMSAT to the tasks of clustering and hash learning, respectively. 
In Section \ref{subsec:approximation}, we discuss an approximation technique for scaling up our method.

\subsection{Review of Regularized Information Maximization for Clustering} \label{subsec:rim}
The RIM \citep{krause2010discriminative} learns a probabilistic classifier $p_{\theta}(y | x)$ such that mutual information \citep{cover2012elements} between inputs and cluster assignments is maximized. At the same time, it regularizes the complexity of the classifier.
Let $X \in \mathcal{X}$ and $Y \in \mathcal{Y} \equiv \{ 0, \ldots, K-1 \}$ denote random variables for data and cluster assignments, respectively, where $K$ is the number of clusters.
The RIM \emph{minimizes} the objective:
\begin{align}
 \mathcal{R}(\theta) - \lambda \vector{I}(X; Y), \label{eq:rim}
\end{align}
where $ \mathcal{R}(\theta)$ is the regularization penalty, and $\vector{I}(X; Y)$ is mutual information between $X$ and $Y$, which depends on $\theta$ through the classifier, $p_{\theta}(y | x)$.
Mutual information measures the statistical dependency between $X$ and $Y$, and is 0 iff they are independent. Hyper-parameter $\lambda \in \mathbb{R}$ trades off the two terms.

\subsection{Information Maximizing Self-Augmented Training} \label{subsec:imsat}
Here, we present two components that make up our IMSAT. We present the Information Maximization part in Section \ref{subsec:framework} and the SAT part in Section \ref{subsec:regl} .
\subsubsection{Information maximization for learning discrete representations} \label{subsec:framework}
We extend the RIM and consider learning $M$-dimensional discrete representations of data. Let the output domain be $\mathcal{Y} = \mathcal{Y}_1 \times \cdots \times \mathcal{Y}_M$, where $\mathcal{Y}_m \equiv \{0, 1, \ldots, V_m - 1 \}, \ 1\leq m \leq M$. 
Let $Y = (Y_1, \ldots, Y_M) \in \mathcal{Y}$ be a random variable for the discrete representation. 
Our goal is to learn a multi-output probabilistic classifier $p_{\theta}(y_1, \ldots, y_M | x)$ that maps similar inputs into similar representations.
For simplicity, we model the conditional probability $p_{\theta}(y_1, \ldots, y_M | x)$ by using the deep neural network depicted in Figure~\ref{fig:image}. Under the model, $\{y_1, \ldots, y_M \}$ are conditionally independent given $x$: 
\begin{align}
p_{\theta}( y_1,  \ldots,  y_M |  x)=  \prod_{m = 1}^{M} p_{\theta}(y_m |  x). \label{eq:condind}
\end{align}

Following the RIM \citep{krause2010discriminative}, we maximize the mutual information between inputs and their discrete representations, while regularizing the multi-output probabilistic classifier.
The resulting objective to \emph{minimize} looks exactly the same as Eq.~\eqref{eq:rim}, except that $Y$ is multi-dimensional in our setting.

\subsubsection{Regularization of deep neural networks via self-augmented training} \label{subsec:regl}
We present an intuitive and flexible regularization objective, termed \emph{Self-Augmented Training (SAT)}. 
SAT uses data augmentation to impose the intended invariance on the data representations. Essentially, SAT penalizes representation dissimilarity between the original data points and augmented ones. 
Let $T: \mathcal{X} \to \mathcal{X}$ denote a pre-defined data augmentation under which the data representations should be invariant. The regularization of SAT made on data point $x$ is
\begin{align}
{\mathcal R}_{{\rm SAT}}(\theta ; x, T(x)) &\nonumber \\
= - \sum_{m = 1}^{M} \sum_{y_m = 0}^{V_m - 1} &  p_{\widehat{\theta}}( y_m | x) \log p_{\theta}(y_m | T(x)), \label{eq:sat_each_data}
\end{align}
where $p_{\widehat{\theta}}( y_m | x)$ is the prediction of original data point $x$, and $\widehat{\theta}$ is the current parameter of the network. 
In Eq.~\eqref{eq:sat_each_data}, the representations of the augmented data are pushed to be close to those of the original data. Since probabilistic classifier $p_{\theta}(y | x)$ is modeled using a deep neural network, it is flexible enough to capture a wide range of invariances specified by the augmentation function $T$. The regularization by SAT is then the average of ${\mathcal R}_{{\rm SAT}}(\theta ; x, T(x))$ over all the training data points:
\begin{align}
{\mathcal R}_{{\rm SAT}}(\theta; T) = \frac{1}{N} \sum_{n = 1}^{N} {\mathcal R}_{{\rm SAT}}(\theta ; x_n, T(x_n)). \label{eq:SAT}
\end{align}

The augmentation function $T$ can either be stochastic or deterministic. It can be designed specifically for the applications of interest. For example, for image data, affine distortion such as rotation, scaling and parallel movement can be used for the augmentation function.

Alternatively, more general augmentation functions that do not depend on specific applications can be considered. A representative example is local perturbations, in which the augmentation function is
\begin{align}
T(x) = x + r, \label{eq:vat}
\end{align}
where $r$ is a small perturbation that does not alter the meaning of the data point. The use of local perturbations in SAT encourages the data representations to be locally invariant. The resulting decision boundaries between different representations tend to lie in low density regions of a data distribution. Such boundaries are generally preferred and follow the low-density separation principle \citep{grandvalet2004semi}.

The two representative regulariztion methods based on local perturbations are: Random Perturbation Training (RPT) \citep{bachman2014learning} and Virtual Adversarial Training (VAT) \citep{miyato2015distributional}.
In RPT, perturbation $r$ is sampled randomly from hyper-sphere $|| r ||_2 = \epsilon$, where $\epsilon$ is a hyper-parameter that controls the range of the local perturbation.
On the other hand, in VAT, perturbation $r$ is chosen to be an \emph{adversarial} direction:
\begin{align}
r = \argmax_{r^{\prime}}\  \{ {\mathcal R}_{{\rm SAT}}(\widehat{\theta} ; x, x + r^{\prime}); \ || r^{\prime} ||_2 \leq \epsilon \}.  \label{eq:adv}
\end{align}
The solution of Eq.~\eqref{eq:adv} can be approximated efficiently by a pair of forward and backward passes. For further details, refer to \citet{miyato2015distributional}.


\subsection{IMSAT for Clustering} 
\label{subsec:adapt_clustering}
In clustering, we can directly apply the RIM \citep{krause2010discriminative} reviewed in Section \ref{subsec:rim}. Unlike the original RIM, however, our method, IMSAT, uses deep neural networks for the classifiers and regularizes them via SAT.
By representing mutual information as the difference between marginal entropy and conditional entropy \citep{cover2012elements}, we have the objective to minimize:
\begin{align}
\mathcal{R}_{{\rm SAT}}(\theta; T) - \lambda \left [ H(Y) - H(Y | X) \right],
\end{align}
where $H(\cdot)$ and $H(\cdot | \cdot)$ are entropy and conditional entropy, respectively. The two entropy terms can be calculated as
\begin{align}
H(Y) &\equiv h(p_\theta(y)) =  h\left(\frac{1}{N}\sum_{i = 1}^N p_{\theta}(y | x) \right), \label{eq:marginal_entropy} \\
H(Y | X) & \equiv \frac{1}{N} \sum_{i = 1}^{N} h(p_{\theta}(y|x_i)) \label{eq:conditional_entropy},
\end{align}
where $h(p(y)) \equiv - \sum_{y^{\prime}} p(y^{\prime}) \log p(y^{\prime})$ is the entropy function.
Increasing the marginal entropy $H(Y)$ encourages the cluster sizes to be uniform, while decreasing the conditional entropy $H(Y | X)$ encourages unambiguous cluster assignments \citep{heading1991unsupervised}.

In practice, we can incorporate our prior knowledge on cluster sizes by modifying $H(Y)$ \citep{krause2010discriminative}. Note that $H(Y) = \log K - {\rm KL}[p_{\theta}(y) ||\  \mathcal{U}]$, where $K$ is the number of clusters, ${\rm KL}[\cdot || \cdot]$ is the Kullback-Leibler divergence, and $\mathcal{U}$ is a uniform distribution. 
Hence, maximization of $H(Y)$ is equivalent to minimization of ${\rm KL}[p_{\theta}(y) ||\  \mathcal{U}]$, which encourages predicted cluster distribution $p_{\theta}(y)$ to be close to $\mathcal{U}$.
\citet{krause2010discriminative} replaced~$\mathcal{U}$ in ${\rm KL}[p_{\theta}(y) ||\  \mathcal{U}]$ with any specified class prior $q(y)$ so that $p_{\theta}(y)$ is encouraged to be close to $q(y)$.
In our preliminary experiments, we found that the resulting $p_{\theta}(y)$ could still be far apart from pre-specified $q(y)$.
To ensure that $p_{\theta}(y)$ is actually close to $q(y)$, we consider the following constrained optimization problem:
\begin{align}
&\min_{\theta}\  \mathcal{R}_{{\rm SAT}}(\theta; T) + \lambda  H(Y | X), \nonumber \\
&\text{subject to}\ \  {\rm KL}[p_{\theta}(y) || \ q(y)] \leq \delta, \label{eq:constraint}
\end{align}
where $\delta > 0$ is a tolerance hyper-parameter that is set sufficiently small so that predicted cluster distribution $p_{\theta}(y)$ is the same as class prior $q(y)$ up to $\delta$-tolerance. 
Eq.~\eqref{eq:constraint} can be solved by using the penalty method \citep{bertsekas1999nonlinear}, which turns the original constrained optimization problem into a series of unconstrained optimization problems.
\ifnum\value{long}<1
Refer to Appendix \ref{app:penalty} of the supplementary material for the detail. 
\else
Refer to Appendix \ref{app:penalty} for the detail. 
\fi

\subsection{IMSAT for Hash Learning} \label{subsec:adapt_hashing}
In hash learning, each data point is mapped into a $D$-bit-binary code. Hence, the original RIM is not directly applicable. Instead, we apply our method for discrete representation learning presented in Section \ref{subsec:framework}.

The computation of mutual information $\vector{I}(Y_1, \ldots, Y_D; X)$, however, is intractable for large $D$ because it involves a summation over an exponential number of terms, each of which corresponds to a different configuration of hash bits.

\citet{brown2009new} showed that mutual information $\vector{I}(Y_1, \ldots, Y_D; X)$ can be expanded as the sum of interaction information \citep{mcgill1954multivariate}:
\begin{align} \label{eq:decomp}
\vector{I}(Y_1, \ldots, Y_D; X) = \sum_{C \subseteq S_Y} \vector{I} (C \cup X), \ \ |C| \geq 1,
\end{align}
where $S_Y \equiv \{ Y_1, \ldots, Y_D\}$. Note that $\vector{I}$ denotes interaction information when its argument is a set of random variables. 
Interaction information is a generalization of mutual information and can take a negative value. 
When the argument is a set of two random variables, the interaction information reduces to mutual information between the two random variables.
Following \citet{brown2009new}, we only retain terms involving pairs of output dimensions in Eq.~\eqref{eq:decomp}, i.e., all terms where $|C| \leq 2$. This gives us
\begin{align}
\sum_{d = 1}^{D} \vector{I}(Y_d; X) + \sum_{1 \leq d \neq d^{\prime} \leq D} \vector{I}(\{Y_d, Y_{d^{\prime}}, X\}).
\end{align}
This approximation ignores the interactions among hash bits beyond the pairwise interactions. 
It is related to the orthogonality constraint that is widely used in the literature to remove redundancy among hash bits \citep{wang2016learning}. 
In fact, the orthogonality constraint encourages the covariance between a pair of hash bits to 0. Thus, it also takes into account the pairwise interactions.

It follows from the definition of interaction information and the conditional independence in Eq.~\eqref{eq:condind} that
\begin{align}
\vector{I}(\{Y_d, Y_{d^{\prime}}, X\}) & \equiv \vector{I}(Y_d; Y_{d^{\prime}} | X) - \vector{I}(Y_d; Y_{d^{\prime}})  \nonumber \\
& =  - \vector{I}(Y_d; Y_{d^{\prime}}).
\end{align}

In summary, our approximated objective to minimize is
\begin{align}
 \mathcal{R}_{{\rm SAT}}(\theta; T) - \lambda \left( \sum_{d = 1}^{D} \vector{I}(X; Y_d) - \sum_{1 \leq d \neq d^{\prime} \leq D} \vector{I}(Y_d; Y_{d^{\prime}}) \right). \label{eq:approx}
\end{align}
The first term regularizes the neural network.
The second term maximizes the mutual information between data and each hash bit, and the third term removes the redundancy among the hash bits.

\subsection{Approximation of the Marginal Distribution}  \label{subsec:approximation}
To scale up our method to large datasets, we would like the objective in Eq.~\eqref{eq:rim} to be amenable to optimization based on mini-batch SGD. 
For the regularization term, we use the SAT in Eq.~\eqref{eq:SAT}, which is the sum of per sample penalties and can be readily adapted to mini-batch computation. For the approximated mutual information in Eq.~\eqref{eq:approx}, we can decompose it into three parts: (i) conditional entropy $H(Y_d | X)$, (ii) marginal entropy $H(Y_d)$, and (iii) mutual information between a pair of output dimensions $\vector{I}(Y_d; Y_{d^{\prime}})$. The conditional entropy only consists of a sum over per example entropies (see Eq.~\eqref{eq:conditional_entropy}); thus, can be easily adapted to mini-batch computation. However, the marginal entropy (see Eq.~\eqref{eq:marginal_entropy}) and the mutual information involve the marginal distribution over a subset of target dimensions, i.e., $p_{\theta}(c) \equiv \frac{1}{N}\sum_{n = 1}^{N} p_{\theta}(c | x_n)$, where $c \subseteq \{ y_1, \ldots, y_M \}$. Hence, the marginal distribution can only be calculated using the entire dataset and is not amenable to the mini-batch setting. Following \citet{springenberg2015unsupervised}, we approximate the marginal distributions using mini-batch data:
\begin{align}
p_{\theta}(c) \approx \frac{1}{|\mathcal{B}|}\sum_{x \in \mathcal{B}} p_{\theta}(c | x) \equiv \widehat{p_{\theta}}^{(\mathcal{B})}(c),
\end{align}
where $\mathcal{B}$ is a set of data in the mini-batch. In the case of clustering, the approximated objective that we actually minimize is an upper bound of the exact objective that we try to minimize.
\ifnum\value{long}<1
Refer to Appendix \ref{sec:mini-batch} of the supplementary material for the detailed discussion.
\else
Refer to Appendix \ref{sec:mini-batch} for the detailed discussion.
\fi


\section{Experiments}\label{sec:experiment}
In this section, we evaluate IMSAT for clustering and hash learning using benchmark datasets.

\subsection{Implementation}
In unsupervised learning, it is not straightforward to determine hyper-parameters by cross-validation. 
Therefore, in all the experiments with benchmark datasets, we used commonly reported parameter values for deep neural networks and avoided dataset-specific tuning as much as possible.
Specifically, inspired by \citet{hinton2012improving}, we set the network dimensionality to $d$-1200-1200-$M$ for clustering across all the datasets, where $d$ and $M$ are input and output dimensionality, respectively. 
For hash learning, we used smaller network sizes to ensure fast computation of mapping data into hash codes.
We used rectified linear units \citep{jarrett2009best, nair2010rectified, glorot2011deep} for all the hidden activations and applied batch normalization \citep{ioffe2015batch} to each layer to accelerate training. 
For the output layer, we used the softmax for clustering and the sigmoids for hash learning. 
Regarding optimization, we used Adam \citep{kingma2014adam} with the step size 0.002.
\ifnum\value{long}<1
Refer to Appendix \ref{app:implementation} of the supplemental material for further details.
\else
Refer to Appendix \ref{app:implementation} for further details.
\fi
Our implementation based on Chainer \citep{tokui2015chainer} is available at {\tt \url{https://github.com/weihua916/imsat}}.

\subsection{Clustering} \label{subsec:clustering}
\subsubsection{Datasets and compared methods}
We evaluated our method for clustering presented in Section \ref{subsec:adapt_clustering} on eight benchmark datasets.
We performed experiments with two variants of the RIM and three variants of IMSAT, each of which uses different classifiers and regularization. 
Table \ref{table:variant} summarizes these variants.
We also compared our IMSAT with existing clustering methods including $K$-means, DEC \citep{xie2016unsupervised}, denoising Auto-Encoder~(dAE)+$K$-means \citep{xie2016unsupervised}. 

\begin{table}[t] 
\caption{Summary of the variants.} \label{table:variant}
\footnotesize
\begin{center}
\begin{tabular}{|l||c|c|c|}\hline
Method & Used classifier & Regularization \\ \hline \hline
Linear RIM  &  Linear & Weight-decay  \\ \hline
Deep RIM & Deep neural nets  & Weight-decay \\ \hline
Linear IMSAT (VAT) & Linear & VAT \\ \hline
{\bf IMSAT (RPT)} & Deep neural nets & RPT \\ \hline
{\bf IMSAT (VAT)} & Deep neural nets & VAT \\ \hline
\end{tabular}
\label{table:clustering_omniglot}
\vspace*{-\baselineskip}
\end{center}
\end{table}

\begin{table*}[t]
\begin{center} 
\caption{Summary of dataset statistics.}
\begin{tabular}{|l||c|c|c|c|}\hline
\label{table:statistics}
Dataset & \#Points & \#Classes & Dimension & \%Largest class \\ \hline \hline
MNIST \citep{lecun1998gradient} & 70000  & 10 & 784 & 11\%  \\ \hline
Omniglot \citep{lake2011one} & 40000  & 100 & 441& 1\%\\ \hline
STL \citep{coates2010analysis}   & 13000 & 10 & 2048 & 10\%    \\  \hline
CIFAR10 \citep{torralba200880} & 60000  &10 &  2048 & 10\%   \\ \hline
CIFAR100  \citep{torralba200880} & 60000  &  100 & 2048 & 1\% \\ \hline
SVHN \citep{netzer2011reading} & 99289 &  10 & 960 & 19\% \\ \hline
Reuters \citep{lewis2004rcv1} & 10000 &  4 & 2000 & 43\% \\ \hline
20news \citep{Lang95} &18040 & 20 & 2000 & 5\% \\ \hline
\end{tabular}
\end{center} 
\end{table*}

\begin{table*}[t] 
\footnotesize
\begin{center}
\caption{Comparison of clustering accuracy on eight benchmark datasets (\%). Averages and standard deviations over twelve trials were reported. Results marked with $\dagger$ were excerpted from \citet{xie2016unsupervised}.}
\begin{tabular}{|l||c|c|c|c|c|c|c|c|c|}\hline
Method & MNIST & Omniglot& STL & CIFAR10 & CIFAR100 & SVHN & Reuters & 20news  \\ \hline \hline
$K$-means & 53.2  & 12.0  & 85.6 & 34.4  & 21.5  & 17.9  & 54.1  & 15.5   \\ \hline
dAE+$K$-means & 79.8 $^{\dagger}$  & 14.1  & 72.2 & 44.2  & 20.8  & 17.4  & 67.2  & 22.1    \\ \hline
DEC  &84.3 $^{\dagger}$ & 5.7 (0.3)  & 78.1 (0.1) &  {\bf 46.9 (0.9)}  & 14.3 (0.6)  & 11.9 (0.4)  &67.3 (0.2)  & 30.8 (1.8)   \\  \hline
Linear RIM  & 59.6 (2.3)  & 11.1 (0.2) &  73.5 (6.5)  &  40.3 (2.1)  &  23.7 (0.8)  & 20.2 (1.4)   &  62.8 (7.8)  & {\bf 50.9 (3.1)}   \\ \hline
Deep RIM  & 58.5 (3.5)   & 5.8 (2.2)  & 92.5 (2.2)  & 40.3 (3.5)  &  13.4 (1.2) & 26.8 (3.2)  & 62.3 (3.9)  & 25.1 (2.8)    \\ \hline
Linear IMSAT (VAT) & 61.1 (1.9)   & 12.3 (0.2)  & 91.7 (0.5)  & 40.7 (0.6)  &  23.9 (0.4) & 18.2 (1.9)  & 42.9 (0.8)  & 43.9 (3.3)    \\ \hline
{\bf IMSAT (RPT)} & 89.6 (5.4)   & 16.4 (3.1)  & 92.8 (2.5)  & 45.5 (2.9)  &  24.7 (0.5) & 35.9 (4.3)  & {\bf 71.9 (6.5)}  & 24.4 (4.7)    \\ \hline
{\bf IMSAT (VAT)} & {\bf 98.4 (0.4)} &  {\bf 24.0 (0.9)}  & {\bf 94.1 (0.4)}  & 45.6 (0.8) & {\bf 27.5 (0.4)}  & {\bf 57.3 (3.9)}  & {\bf 71.0 (4.9)}  & 31.1 (1.9)   \\ \hline
\end{tabular}
\label{table:clustering}
\end{center}
\end{table*}

A brief summary of dataset statistics is given in Table \ref{table:statistics}.
In the experiments, our goal was to discover clusters that correspond well with the ground-truth categories.
For the STL, CIFAR10 and CIFAR100 datasets, raw pixels are not suited for our goal because color information is dominant.
We therefore applied 50-layer pre-trained deep residual networks \citep{he2016deep} to extract features and used them for clustering. Note that since the residual network was trained on ImageNet, each class of the STL dataset (which is a subset of ImageNet) was expected to be well-separated in the feature space. 
For Omniglot, 100 types of characters were sampled, each containing 20 data points. Each data point was augmented 20 times by the stochastic affine distortion described in Appendix \ref{app:augmentation}.
For SVHN, each image was represented as a 960-dimensional GIST feature \citep{oliva2001modeling}.
For Reuters and 20news, we removed stop words and retained the 2000 most frequent words. We then used {\it tf-idf} features.
\ifnum\value{long}<1
Refer to Appendix \ref{app:dataset} of the supplementary material for further details. 
\else
Refer to Appendix \ref{app:dataset} for further details.
\fi

\subsubsection{Evaluation metric}
Following \citet{xie2016unsupervised}, we set the number of clusters to the number of ground-truth categories and evaluated clustering performance with unsupervised clustering accuracy (ACC):  
\begin{align}
{\rm ACC} = \max_{m} \frac{\sum_{n = 1}^{N} {\bf 1}\{ l_n = m(c_n)\}}{N},
\end{align}
where $l_n$ and $c_n$ are the ground-truth label and cluster assignment produced using the algorithm for $x_n$, respectively. The $m$ ranges over all possible one-to-one mappings between clusters and labels. The best mapping can be efficiently computed using the Hungarian algorithm \citep{kuhn1955hungarian}.

\subsubsection{Hyper-parameter selection} \label{sec:hyper_clustering}

In unsupervised learning, it is not straightforward to determine hyper-parameters by cross-validation. 
Hence, we fixed hyper-parameters across all the datasets unless there was an objective way to select them. 
For $K$-means, we tried 12 different initializations and reported the results with the best objectives.
For dAE+$K$-means and DEC \citep{xie2016unsupervised}, we used the recommended hyper-parameters for the network dimensionality and annealing speed.

Inspired by the automatic kernel width selection in spectral clustering \citep{zelnik2004self}, we set the perturbation range, $\epsilon$, on data point $x$ in VAT and RPT as
\begin{align}
\epsilon(x) = \alpha \cdot \sigma_t(x),
\end{align}
where $\alpha$ is a scalar and $\sigma_t(x)$ is the Euclidian distance to the $t$-th neighbor of $x$. In our experiments, we fixed $t = 10$. For Linear IMSAT (VAT), IMSAT (RPT) and IMSAT (VAT), we fixed $\alpha = 0.4, 2.5$ and $0.25$, respectively, which performed well across the datasets.

For the methods shown in Table \ref{table:variant}, we varied one hyper-parameter and chose the best one that performed well across the datasets. 
More specifically, for Linear RIM and Deep RIM, we varied the decay rate over $0.0025 \cdot 2^{i}, i = 0, 1, \ldots, 7$. For the three variants of IMSAT, we varied $\lambda$ in Eq.~\eqref{eq:unconstrained} for $0.025 \cdot 2^{i}, i = 0, 1, \ldots, 7$. 
We set $q$ to be the uniform distribution and let $\delta = 0.01 \cdot h(q(y))$ in Eq.~\eqref{eq:constraint} for the all experiments. 

Consequently, we chose 0.005 for decay rates in both Linear RIM and Deep RIM. Also, we set $\lambda = 1.6, 0.05$ and $0.1$ for Linear IMSAT (VAT), IMSAT (RPT) and IMSAT (VAT), respectively. We hereforth fixed these hyper-parameters throughout the experiments for both clustering and hash learning. 
\ifnum\value{long}<1
In Appendix \ref{app:hyper} of the supplementary material, we report all the experimental results and the criteria to choose the parameters.
\else
In Appendix \ref{app:hyper}, we report all the experimental results and the criteria to choose the parameters.
\fi

\begin{figure*}[t]
\begin{minipage}[t]{0.5\hsize}
\begin{center}
\centerline{\includegraphics[width=8.25cm, clip]{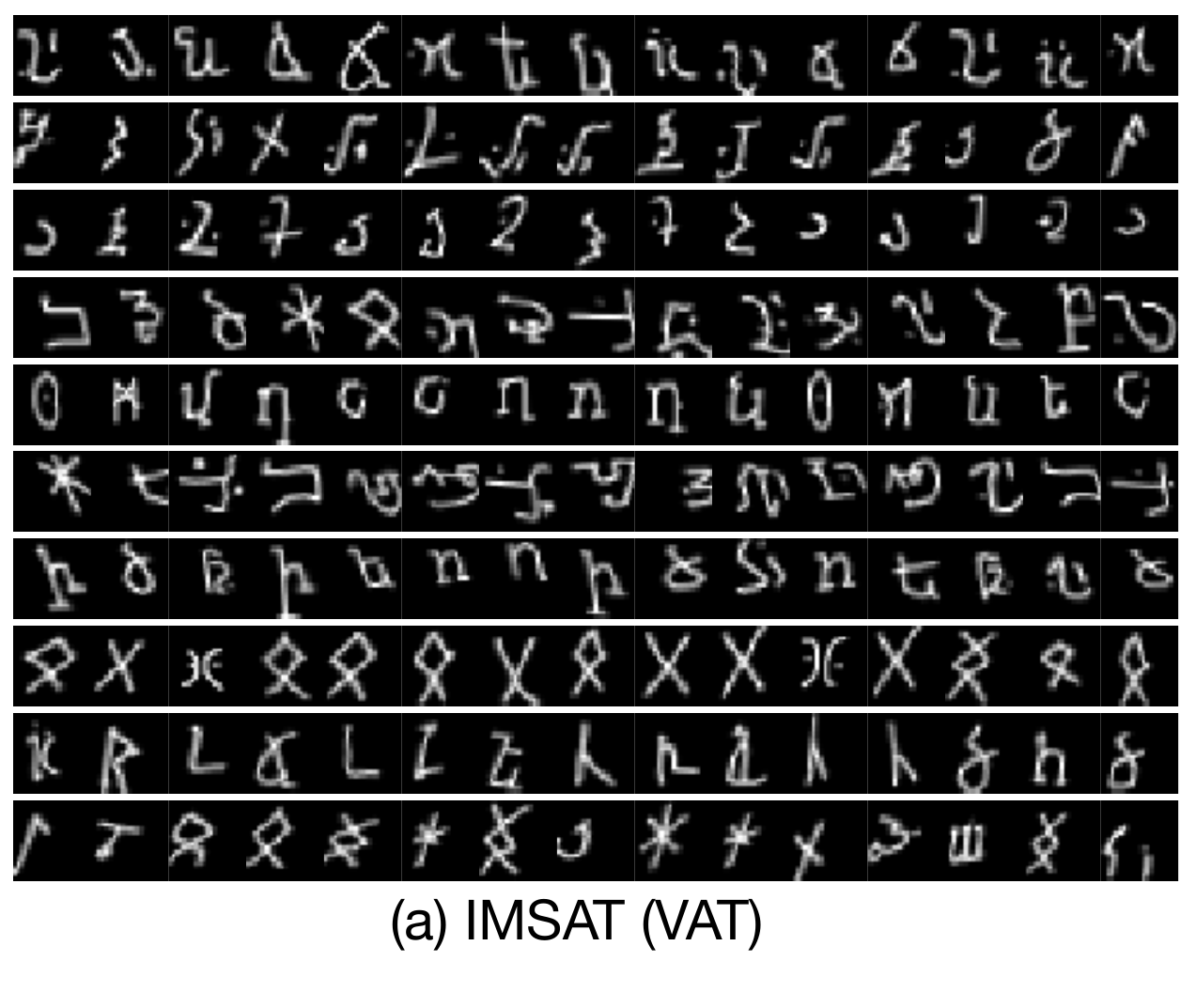}}
\label{image}
\end{center}
\end{minipage}
\begin{minipage}[t]{0.5\hsize}
\begin{center}
\centerline{\includegraphics[width=8.25cm, clip]{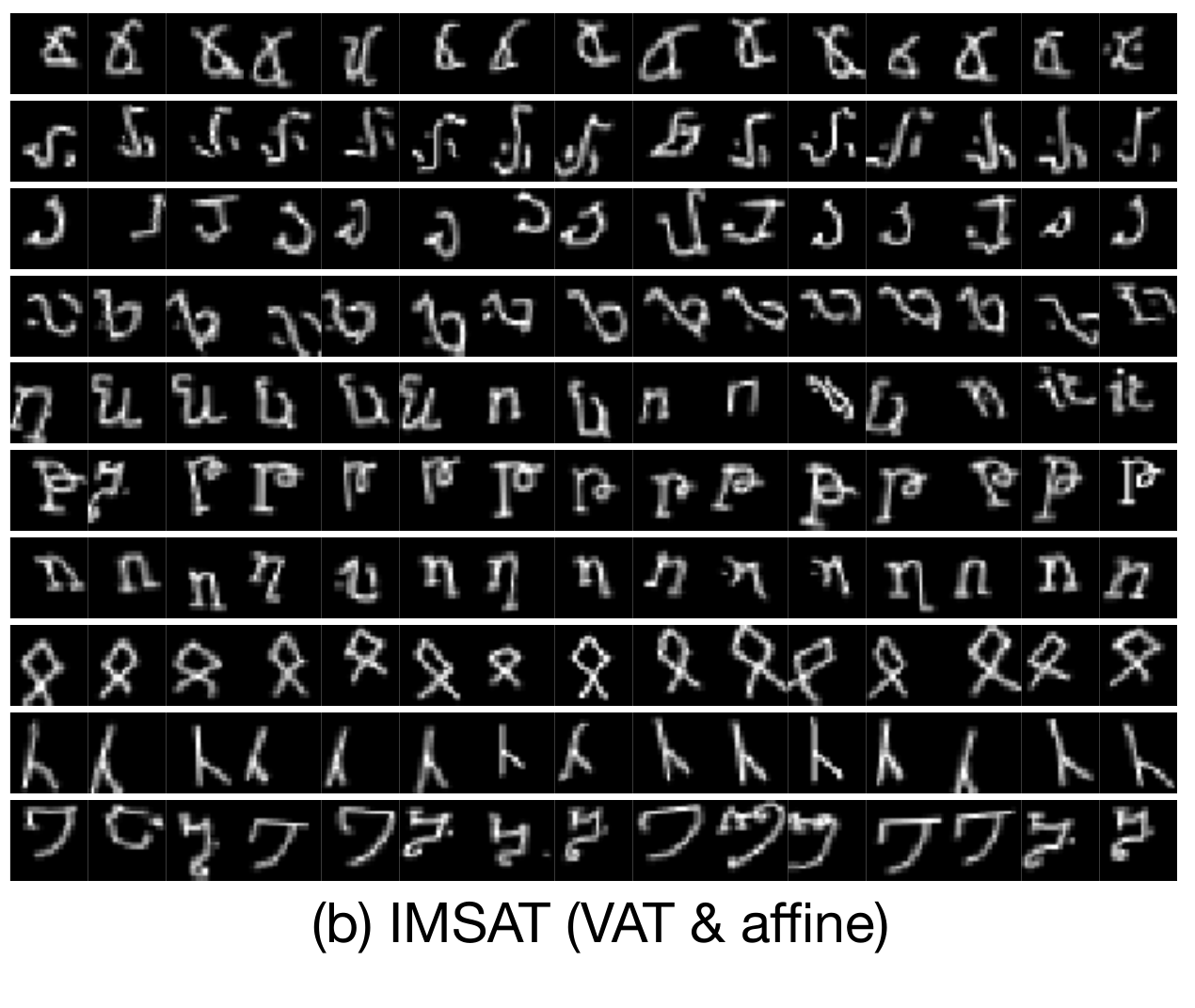}}
\label{image}
\end{center}
\end{minipage}
\vspace{-1cm}
\caption{Randomly sampled clusters of Omniglot discovered using (a) IMSAT (VAT) and (b) IMSAT (VAT \& affine). Each row contains randomly sampled data points in same cluster.}\label{fig:omniglot}
\end{figure*} 

\begin{table}[t] 
\caption{Comparison of clustering accuracy on the Omniglot dataset using IMSAT with different types of SAT.}
\begin{center}
\begin{tabular}{|l||c|c|}\hline
Method & Omniglot \\ \hline \hline
IMSAT (VAT)  &  24.0 (0.9)   \\ \hline
IMSAT (affine) & 45.1 (2.0) \\ \hline
IMSAT (VAT \& affine) & {\bf 70.0 (2.0)}  \\ \hline
\end{tabular}
\vspace*{-\baselineskip}
\label{table:clustering_omniglot}
\end{center}
\end{table}


\subsubsection{Experimental results } \label{sec:exp_clustering}

In Table \ref{table:clustering}, we compare clustering performance across eight benchmark datasets. 
We see that IMSAT (VAT) performed well across the datasets. 
The fact that our IMSAT outperformed Linear RIM, Deep RIM and Linear IMSAT (VAT) for most datasets suggests the effectiveness of using deep neural networks with an end-to-end regularization via SAT. 
Linear IMSAT (VAT) did not perform well even with the end-to-end regularization probably because the linear classifier was not flexible enough to model the intended invariance of the representations.
We also see from Table \ref{table:clustering} that IMSAT (VAT) consistently outperformed IMSAT (RPT) in our experiments.  
This suggests that VAT is an effective regularization method in \emph{unsupervised} learning scenarios.

We further conducted experiments on the Omniglot dataset to demonstrate that clustering performance can be improved by incorporating domain-specific knowledge in the augmentation function of SAT. Specifically, we used the affine distortion in addition to VAT for the augmented function of SAT. We compared the clustering accuracy of IMSAT with three different augmentation functions: VAT, affine distortion, and the combination of VAT \& affine distortion, in which we simply set the regularization to be
\begin{align}
\frac{1}{2} \cdot \mathcal{R}_{{\rm SAT}}(\theta; T_{{\rm VAT}}) + \frac{1}{2} \cdot \mathcal{R}_{{\rm SAT}}(\theta; T_{{\rm affine}}),
\end{align}
where $T_{{\rm VAT}}$ and $T_{{\rm affine}}$ are augmentation functions of VAT and affine distortion, respectively. For $T_{{\rm affine}}$, we used the stochastic affine distortion function defined 
\ifnum\value{long}<1
in Appendix \ref{app:augmentation} of the supplementary material.
\else
in Appendix \ref{app:augmentation}. 
\fi

We report the clustering accuracy of Omniglot in Table~\ref{table:clustering_omniglot}. 
We see that including affine distortion in data augmentation significantly improved clustering accuracy.
Figure~\ref{fig:omniglot} shows ten randomly selected clusters of the Omniglot dataset that were found using IMSAT (VAT) and IMSAT (VAT \& affine distortion).
We observe that IMSAT (VAT \& affine distortion) was able to discover cluster assignments that are invariant to affine distortion as we intended. 
These results suggest that our method successfully captured the invariance in the hand-written character recognition in an unsupervised way.



\subsection{Hash Learning} \label{subsec:hashing}

\subsubsection{Datasets and Compared methods}
We evaluated our method for hash learning presented in Section \ref{subsec:adapt_hashing} on two benchmark datasets: MNIST and CIFAR10 datasets. Each data sample of CIFAR10 is represented as a 512-dimensional GIST feature \citep{oliva2001modeling}. 
Our method was compared against several unsupervised hash learning methods: spectral hashing \citep{weiss2009spectral}, PCA-ITQ \citep{gong2013iterative}, and Deep Hash \citep{erin2015deep}. 
We also compared our method to the hash versions of Linear RIM and Deep RIM. 
For our IMSAT, we used VAT for the regularization.
We used the same hyper-parameters as in Section \ref{sec:hyper_clustering}. 

\subsubsection{Evaluation metric}
Following \citet{erin2015deep}, we used three evaluation metrics to measure the performance of the different methods: 1) mean average precision (mAP); 2) precision at $N=500$ samples; and 3) Hamming look-up result where the hamming radius is set as $r=2$. We used the class labels to define the neighbors.
We repeated the experiments ten times and took the average as the final result. 


\subsubsection{Experimental results} \label{sec:exp_hashing}
The MNIST and CIFAR10 datasets both have 10 classes, and contain 70000 and 60000 data points, respectively. 
Following \citet{erin2015deep}, we randomly sampled 1000 samples, 100 per class, as the query data and used the remaining data as the gallery set. 

\begin{table*}[t] 
\begin{center}
\caption{Comparison of hash performance for 16-bit hash codes (\%). Averages and standard deviations over ten trials were reported. Experimental results of Deep Hash and the previous methods were excerpted from \citet{erin2015deep}.}
\begin{tabular}{|l||c|c|c|c|c|c|c|}\hline
Method & \multicolumn{2}{|c|}{Hamming ranking (mAP)} &\multicolumn{2}{|c|}{precision @ sample = 500} & \multicolumn{2}{|c|}{precision @ r = 2} \\ \cline{2-7}
(Dimensions of hidden layers)             & MNIST & CIFAR10 & MNIST & CIFAR10 & MNIST & CIFAR10 \\ \hline
Spectral hash \citep{weiss2009spectral} & 26.6  &12.6  & 56.3 & 18.8  & 57.5  & 18.5  \\ \hline
PCA-ITQ \citep{gong2013iterative} & 41.2   & 15.7  & 66.4 & 22.5  & 65.7 &  22.6  \\ \hline
Deep Hash (60-30) &43.1  & 16.2  & 67.9 &  23.8  & 66.1  & 23.3   \\  \hline
Linear RIM  & 35.9 (0.6) & {\bf 24.0 (3.5)}  & 68.9 (1.1) &15.9 (0.5)  &71.3 (0.9)   & 14.2 (0.3)    \\ \hline
Deep RIM (60-30) & 42.7 (2.8)  & 15.2 (0.5)  & 67.9 (2.7)   &  21.8 (0.9)  &  65.9 (2.7)  & 21.2 (0.9)   \\ \hline
Deep RIM (200-200) &43.7 (3.7)  & 15.6 (0.6) &  68.7 (4.9) & 21.6 (1.2)  &  67.0 (4.9) &  21.1 (1.1)   \\ \hline
Deep RIM (400-400) &43.9 (2.7)  & 15.4 (0.2) &  69.0 (3.2) &  21.5 (0.4)  &  66.7 (3.2) & 20.9 (0.3)  \\ \hline
{\bf IMSAT (VAT)} (60-30)  & 61.2 (2.5) & 19.8 (1.2)  & 78.6 (2.1) & 21.0 (1.8)  & 76.5 (2.3)   &19.3 (1.6)      \\ \hline
{\bf IMSAT (VAT)} (200-200)  & 80.7 (2.2) &  21.2 (0.8)  & 95.8 (1.0) & {\bf 27.3 (1.3)}  & 94.6 (1.4)   & 26.1 (1.3)    \\ \hline
{\bf IMSAT (VAT)} (400-400)  & {\bf 83.9 (2.3)} & 21.4 (0.5)  & {\bf 97.0 (0.8)} & {\bf 27.3 (1.1)}   & {\bf 96.2 (1.1)}   &  {\bf 26.4 (1.0)}     \\ \hline
\end{tabular}
\label{table:hash_16}
\end{center}
\end{table*}

We tested performance for 16 and 32-bit hash codes.
In practice, fast computation of hash codes is crucial for fast information retrieval. Hence, small networks are preferable.
We therefore tested our method on three different network sizes: the same ones as Deep Hash \citep{erin2015deep}, $d$-200-200-$M$, and $d$-400-400-$M$. Note that Deep Hash used $d$-60-30-$M$ and $d$-80-50-$M$ for learning 16 and 32-bit hash codes, respectively.

Table \ref{table:hash_16} lists the results for 16-bit hash.
Due to the space constraint, we report the results for 32-bit hash codes 
\ifnum\value{long}<1
in Appendix \ref{app:exp_hash32} of the supplementary material,
\else
in Appendix \ref{app:exp_hash32},
\fi
but the results showed a similar tendency as that of 16-bit hash codes. We see from Table \ref{table:hash_16} that IMSAT with the largest network sizes (400-400) achieved competitive performance in both datasets. The performance of IMSAT improved significantly when slightly bigger networks (200-200) were used, while the performance of Deep RIM did not improve much with the larger networks. We deduce that this is because we can better model the local invariance by using more flexible networks. 
Deep RIM, on the other hand, did not significantly benefit from the larger networks, because the additional flexibility of the networks was not used by the global function regularization via weight-decay.\footnote{Hence, we deduce that Deep Hash, which is only regularized by weight-decay, would not benefit much by using larger networks.} 
\ifnum\value{long}<1
In Appendix \ref{app:hash_toy} of the supplementary material, our deduction is supported using a toy dataset.
\else
In Appendix \ref{app:hash_toy}, our deduction is supported using a toy dataset.
\fi

\vspace{-0.2cm}
\section{Conclusion \& Future Work}\label{sec:conclusion}
In this paper, we presented IMSAT, an information-theoretic method for unsupervised discrete representation learning using deep neural networks. 
Through extensive experiments, we showed that intended discrete representations can be obtained by directly imposing the invariance to data augmentation on the prediction of neural networks in an end-to-end fashion. For future work, it is interesting to apply our method to structured data, i.e., graph or sequential data, by considering appropriate data augmentation.
\vspace{-0.2cm}
\section*{Acknowledgements} 
We thank Brian Vogel for helpful discussions and insightful reviews on the paper. This paper is based on results obtained from Hu's internship at Preferred Networks, Inc.

\nocite{langley00}

\bibliography{reference}
\bibliographystyle{icml2016}

\clearpage
\onecolumn
\appendix
\section{Relation to Denoising and Contractive Auto-encoders} \label{app:dae}
Our method is related to denoising auto-encoders \citep{vincent2008extracting}. Auto-encoders maximize a lower bound of mutual information \citep{cover2012elements} between inputs and their hidden representations \citep{vincent2008extracting}, while the denoising mechanism regularizes the auto-encoders to be locally invariant. However, such a regularization does not necessarily impose the invariance on the \emph{hidden representations} because the decoder network also has the flexibility to model the invariance to data perturbations. SAT is more direct in imposing the intended invariance on hidden representations predicted by the encoder network.

Contractive auto-encoders \citep{rifai2011contractive} directly impose the local invariance on the encoder network by minimizing the Frobenius norm of the Jacobian with respect to the weight matrices. However, it is empirically shown that such regularization attained lower generalization performance in supervised and semi-supervised settings than VAT, which regularizes neural networks in an end-to-end fashion \citep{miyato2015distributional}. Hence, we adopted the end-to-end regularization in our unsupervised learning. In addition, our regularization, SAT, has the flexibility of modeling other types invariance such as invariance to affine distortion, which cannot be modeled with the contractive regularization. Finally, compared with the auto-encoders approaches, our method does not require learning the decoder network; thus, is computationally more efficient.  

\section{Penalty Method and its Implementation} \label{app:penalty}
Our goal is to optimize the constrained objective of Eq.~\eqref{eq:constraint}:
\begin{align}
&\min_{\theta}\  \mathcal{R}_{{\rm SAT}}(\theta; T) + \lambda  H(Y | X), \nonumber \\
&\text{subject to}\ \  {\rm KL}[p_{\theta}(y) || \ q(y)] \leq \delta. \nonumber
\end{align}
We use the penalty method \citep{bertsekas1999nonlinear} to solve the optimization.
We introduce a scalar parameter $\mu$ and consider minimizing the following unconstrained objective:
\begin{align}
\mathcal{R}_{{\rm SAT}}(\theta; T) + \lambda  H(Y | X) + \mu \max \{{\rm KL}[p_{\theta}(y) || \ q(y)] - \delta, 0 \}. \label{eq:unconstrained}
\end{align}
We gradually increase $\mu$ and solve the optimization of Eq.~\eqref{eq:unconstrained} for a fixed $\mu$. Let $\mu^{\ast}$ be the smallest value for which the solution of Eq.~\eqref{eq:unconstrained} satisfies the constraint of Eq.~\eqref{eq:constraint}. The penalty method ensures that the solution obtained by solving Eq.~\eqref{eq:unconstrained} with $\mu = \mu^{\ast}$ is the same as that of the constrained optimization of Eq.~\eqref{eq:constraint}.

In experiments in Section \ref{subsec:clustering}, we increased $\mu$ in the order of $\lambda, 2\lambda, 4\lambda, 6\lambda, \ldots$ until the solution of Eq.~\eqref{eq:unconstrained} satisfied the constraint of Eq.~\eqref{eq:constraint}.

\section{On the Mini-batch Approximation of theMmarginal Distribution} \label{sec:mini-batch}
The mini-batch approximation can be validated for the clustering scenario in Eq.~\eqref{eq:constraint} as follows. 
By the convexity of the KL divergence \cite{cover2012elements} and Jensen's inequality, we have
\begin{align}
\mathbb{E}_{\mathcal{B}}[{\rm KL}[\widehat{p_{\theta}}^{(\mathcal{B})}(y) || q(y)] ] \geq {\rm KL}[p_{\theta}(y) || q(y)] \geq 0,  \label{eq:aa}
\end{align}
where the first expectation is taken with respect to the randomness of the mini-batch selection. Therefore, in the penalty method, the constraint on the exact KL divergence, i.e., ${\rm KL}[p_{\theta}(y) || \ q(y)] \leq \delta$ can be satisfied by minimizing its upper bound, which is the approximated KL divergence $\mathbb{E}_{\mathcal{B}}[{\rm KL}[\widehat{p_{\theta}}^{(\mathcal{B})}(y) || q(y)] ]$. Obviously, the approximated KL divergence is amenable to the mini-batch setting; thus, can be minimized with SGD.

\section{Implementation Detail} \label{app:implementation}
We set the size of mini-batch to 250, and ran 50 epochs for each dataset. 
We initialized weights following \citet{he2015delving}: each element of the weight is initialized by the value drawn independently from Gaussian distribution whose mean is 0, and standard deviation is $scale \times \sqrt{2/fan_{in}}$, where $fan_{in}$ is the number of input units. 
We set the $scale$ to be 0.1-0.1-0.0001 for weight matrices from the input to the output. The bias terms were all initialized with 0.

\section{Datasets Description} \label{app:dataset}
\begin{itemize}
\item {\bf MNIST}: A dataset of hand-written digit classification \citep{lecun1998gradient}. The value of each pixel was transformed linearly into an interval [-1, 1].
\item {\bf Omniglot}: A dataset of hand-written character recognition \citep{lake2011one}, containing examples from 50 alphabets ranging from well-established international languages. We sampled 100 types of characters from four alphabets, Magi, Anglo-Saxon Futhorc, Arcadian, and Armenian. Each character contains 20 data points. Since the original data have high resolution (105-by-105 pixels), each data point was down-sampled to 21-by-21 pixels. We also augmented each data point 20 times by thestochastic affine distortion explained in Appendix \ref{app:augmentation}.
\item {\bf STL}: A dataset of 96-by-96 color images acquired from labeled examples on ImageNet \citep{coates2010analysis}. Features were extracted using 50-layer pre-trained deep residual networks \citep{he2016deep} available online as a caffe model. Note that since the residual network is also trained on ImageNet, we expect that each class is separated well in the feature space.
\item {\bf CIFAR10}: A dataset of 32-by-32 color images with ten object classes, which are from the Tiny image dataset \citep{torralba200880}. Features were extracted using the 50-layer pre-trained deep residual networks \citep{he2016deep}.
\item {\bf CIFAR100}:  A dataset 32-by-32 color images with 100 refined object classes, which are from the Tiny image dataset \citep{torralba200880}. Features were extracted using the 50-layer pre-trained deep residual networks \citep{he2016deep}.
\item {\bf SVHN}: A dataset with street view house numbers \citep{netzer2011reading}. Training and test images were both used. Each image was represented as a 960-dimensional GIST feature \citep{oliva2001modeling}.
\item {\bf Reuters}: A dataset with English news stories labeled with a category tree \citep{lewis2004rcv1}. Following DEC \citep{xie2016unsupervised}, we used four categories: corporate/industrial, government/social, markets, and economics as labels. The preprocessing was the same as that used by \citet{xie2016unsupervised}, except that we removed stop words. As \citet{xie2016unsupervised} did, 10000 documents were randomly sampled, and {\it tf-idf} features were used.
\item {\bf 20news}: A dataset of newsgroup documents, partitioned nearly evenly across 20 different newsgroups\footnote{{\tt \url{http://qwone.com/~jason/20Newsgroups/}}}. As Reuters dataset, stop words were removed, and the 2000 most frequent words were retained. Documents with less than ten words were then removed, and {\it tf-idf} features were used.
\end{itemize}
For the STL, CIFAR10 and CIFAR100 datasets, each image was first resized into a 224-by-224 image before its feature was extracted using the deep residual network.

\section{Affine Distortion for the Omniglot Dataset} \label{app:augmentation}
We applied stochastic affine distortion to data points in Omniglot. The affine distortion is similar to the one used by \citet{koch2015siamese}, except that we applied the affine distortion on down-sampled images in our experiments. The followings are the stochastic components of the affine distortion used in our experiments. Our implementation of the affine distortion is based on scikit-image\footnote{http://scikit-image.org/}.
\begin{itemize}
\item Random scaling along $x$ and $y$-axis by a factor of $(s_x, s_y)$, where $s_x$ and $s_y$ are drawn uniformly from interval $[0.8, 1.2].$
\item Random translation along $x$ and $y$-axis by $(t_x, t_y)$, where $t_x$ and $t_y$ are drawn uniformly from interval $[-0.4, 0.4].$
\item Random rotation by $\theta$, where $\theta$ is drawn uniformly from interval $[-10^\circ, 10^\circ].$
\item Random shearing along $x$ and $y$-axis by $(\rho_x, \rho_y)$, where $\rho_x$ and $\rho_y$ are drawn uniformly from interval $[-0.3, 0.3].$
\end{itemize}

\begin{figure*}[t]
\begin{minipage}[t]{0.5\hsize}
\begin{center}
\centerline{\includegraphics[width=8cm, clip]{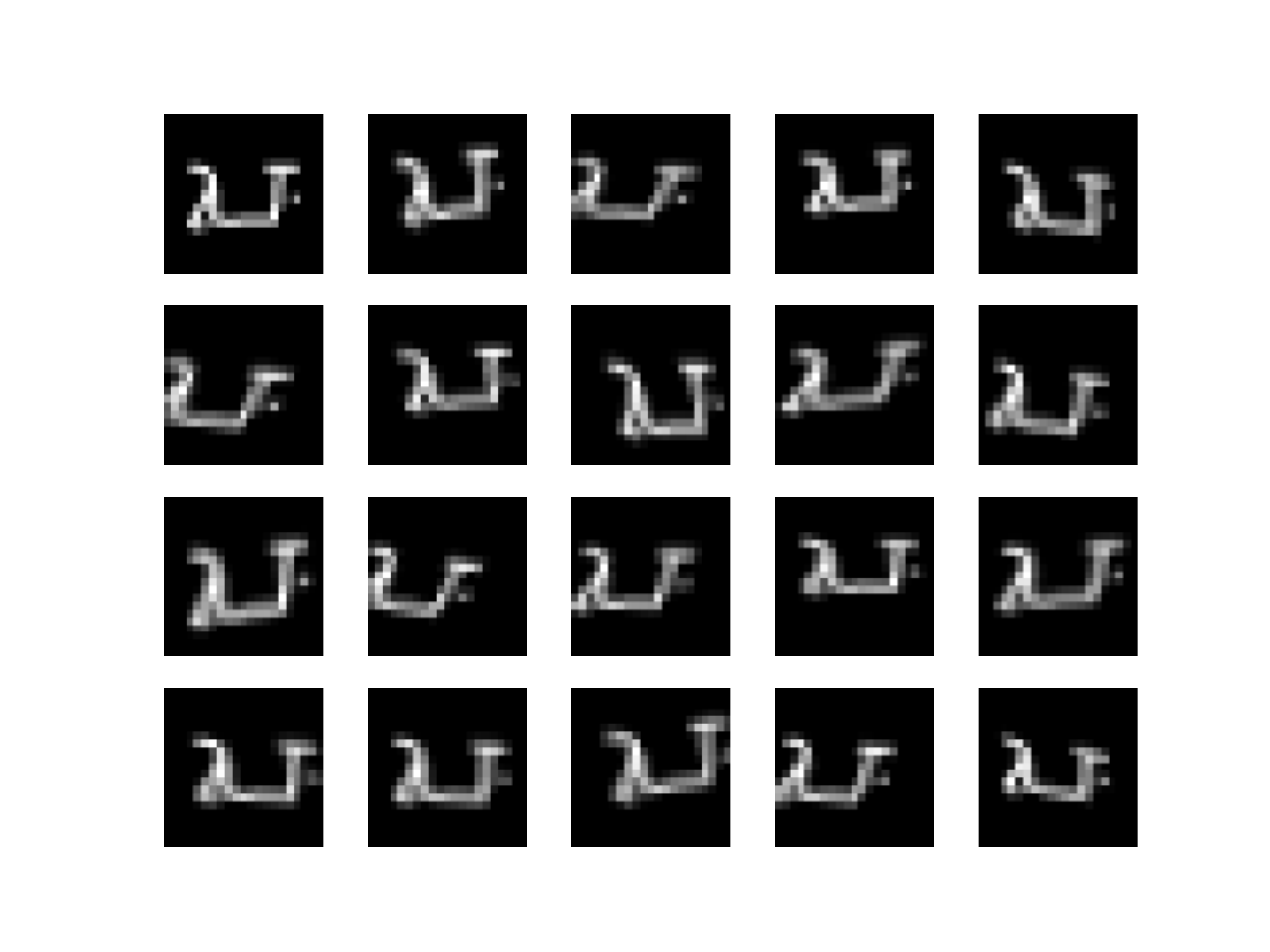}}
\label{image}
\end{center}
\end{minipage}
\begin{minipage}[t]{0.5\hsize}
\begin{center}
\centerline{\includegraphics[width=8cm, clip]{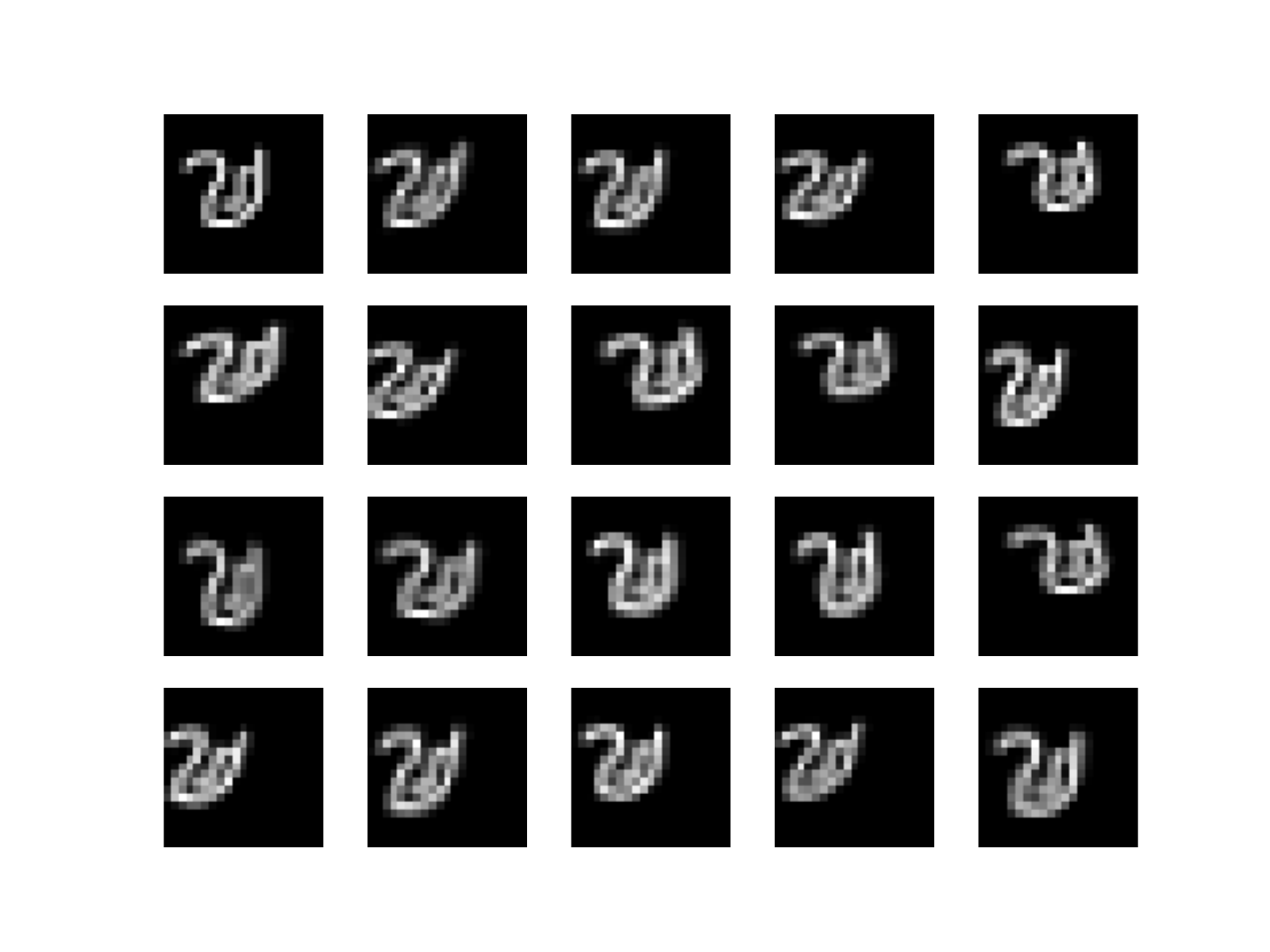}}
\label{image}
\end{center}
\end{minipage}
\vspace{-1cm}
\caption{Examples of the random affine distortion used in our experiments. Images in the top left side are stochastically transformed using the affine distortion.}\label{fig:omniglot_affine}
\end{figure*}

Figure.~\ref{fig:omniglot_affine} shows examples of the random affine distortion.

\section{Hyper-parameter Selection} \label{app:hyper}
\begin{figure*}[t]
\begin{center}
\centerline{\includegraphics[width=17cm, clip]{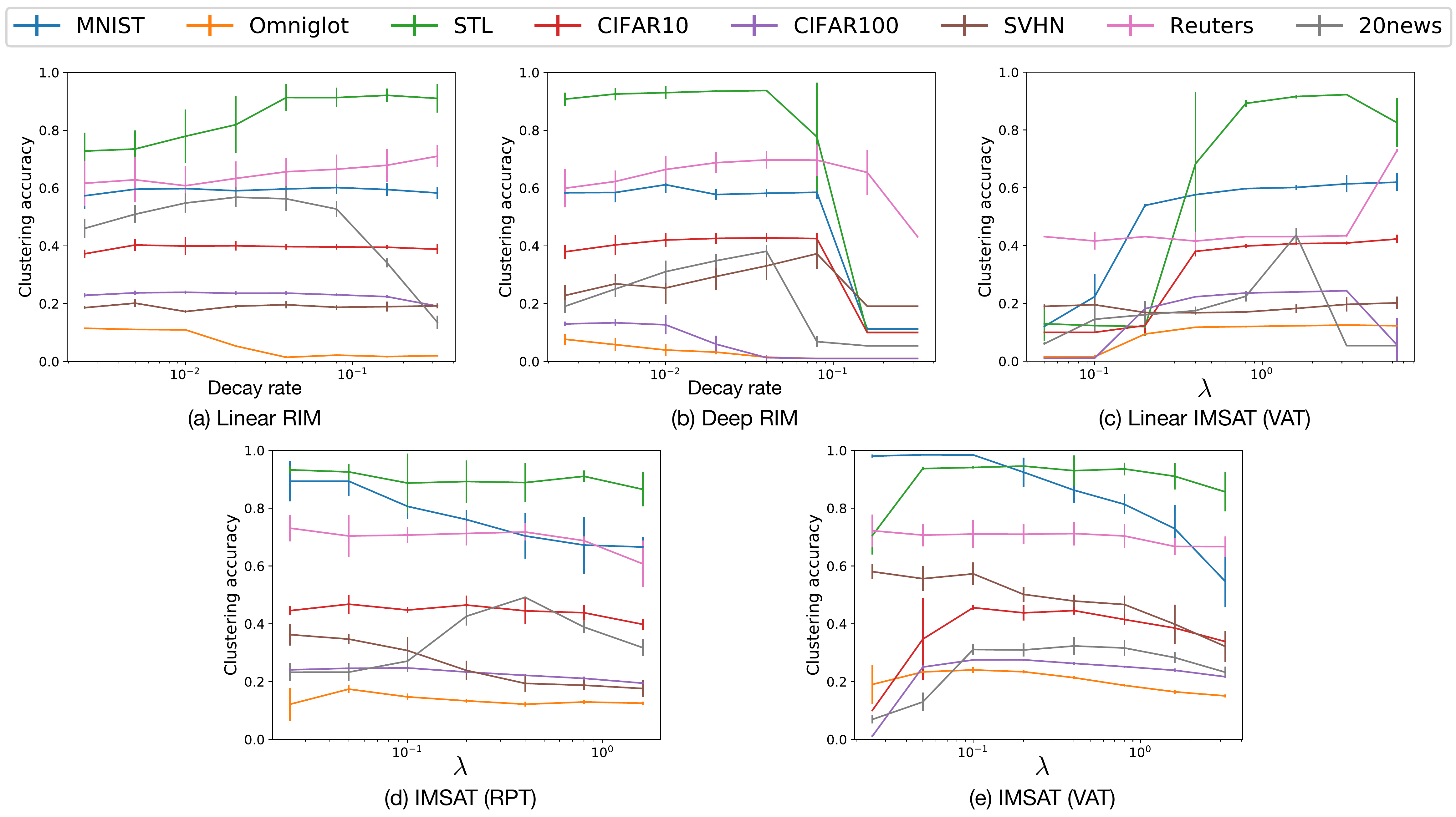}}
\vspace{-0.3cm}
\caption{Relationship between hyper-parameters and clustering accuracy for 8 benchmark datasets with different methods: (a) Linear RIM, (b) Deep RIM, (c) Linear IMSAT (VAT), (d) IMSAT (RPT), and (e) IMSAT (VAT).}\label{fig:rim_comp}
\end{center}
\vspace{-0.8cm}
\end{figure*}

In Figure~\ref{fig:rim_comp} we report the experimental results for different hyper-parameter settings. 
We used Eq.~\eqref{eq:criterion} as a criterion to select hyper-parameter, $\beta^{\ast}$, which performed well across the datasets. 
\begin{align}
\beta^{\ast} = \argmax_{\beta} \sum_{{\rm dataset}} \frac{{\rm ACC}(\beta, {\rm dataset})}{{\rm ACC}(\beta^{\ast}_{\rm dataset}, {\rm dataset})} \label{eq:criterion},
\end{align}
where $\beta^{\ast}_{\rm dataset}$ is the best hyper-parameter for the dataset, and ${\rm ACC}(\beta, {\rm dataset})$ is the clustering accuracy when hyper-parameter $\beta$ is used for the dataset. 
According to the criterion, we set 0.005 for decay rates in both Linear RIM and Deep RIM. Also, we set $\lambda = 1.6, 0.05$ and $0.1$ for Linear IMSAT (VAT), IMSAT (RPT) and IMSAT (VAT), respectively.

\section{Experimental Results on Hash Learning with 32-bit Hash Codes} \label{app:exp_hash32}
Table \ref{table:hash_32} lists the results on hash learning when 32-bit hash codes were used.
We observe that IMSAT with the largest network sizes (400-400) exhibited competitive performance in both datasets. The performance of IMSAT improved significantly when we used slightly larger networks (200-200), while the performance of Deep RIM did not improve much with the larger networks.
\begin{table*}[t] 
\begin{center}
\caption{Comparison of hash performance for 32-bit hash codes (\%). Averages and standard deviations over ten trials were reported. Experimental results of Deep Hash and the previous methods are excerpted from \citet{erin2015deep}.}
\begin{tabular}{|l||c|c|c|c|c|c|c|}\hline
Method & \multicolumn{2}{|c|}{Hamming ranking (mAP)} &\multicolumn{2}{|c|}{precision @ sample = 500} & \multicolumn{2}{|c|}{precision @ r = 2} \\ \cline{2-7}
(Network dimensionality)             & MNIST & CIFAR10 & MNIST & CIFAR10 & MNIST & CIFAR10 \\ \hline
Spectral hash \citep{weiss2009spectral}& 25.7  & 12.4  &61.3 & 19.7  &  65.3 & 20.6  \\ \hline
PCA-ITQ \citep{gong2013iterative} & 43.8   & 16.2  &  74.0  & 25.3  & 73.1  & 15.0  \\ \hline
Deep Hash (80-50)&45.0  & 16.6 & 74.7 &  26.0  & 73.3  & 15.8   \\  \hline
Linear RIM  &29.7 (0.4)   & {\bf 21.2 (3.0)} &   68.9 (0.9)   &  16.7 (0.8)  &  60.9 (2.2)  &  15.2 (0.9)  \\ \hline
Deep RIM (80-50) &34.8 (0.7)   & 14.2 (0.3) &  72.7 (2.2) & 24.0 (0.9)  &  72.6 (2.1) & 23.5 (1.0)   \\ \hline
Deep RIM (200-200) &36.5 (0.8  & 14.1 (0.2) &  76.2 (1.7) & 23.7 (0.7)  &  75.9 (1.6) & 23.3 (0.7)   \\ \hline
Deep RIM (400-400) &37.0 (1.2)  & 14.2 (0.4) & 76.1 (2.2) & 23.9 (1.3)  &  75.7 (2.3)& 23.7 (1.2)   \\ \hline
{\bf IMSAT (VAT)} (80-50)  & 55.4 (1.4) & 20.0 (5.5)  & 87.6 (1.3)& 23.5 (3.4)  & 88.8 (1.3)  & 22.4 (3.2)     \\ \hline
{\bf IMSAT (VAT)} (200-200)  & 62.9 (1.1)& 18.9 (0.7)  &  96.1 (0.6) & 29.8 (1.6)  &  95.8 (0.4)   & {\bf 29.1 (1.4)}     \\ \hline
{\bf IMSAT (VAT)} (400-400)  & {\bf 64.8 (0.8)} & 18.9 (0.5)  & {\bf 97.3 (0.4)} & {\bf 30.8 (1.2)}  &{\bf 96.7 (0.6)}   & {\bf 29.2 (1.2)}     \\ \hline
\end{tabular}
\label{table:hash_32}
\end{center}
\end{table*}

\section{Comparisons of Hash Learning with Different Regularizations and Network Sizes Using Toy Dataset} \label{app:hash_toy}
We used a toy dataset to illustrate that IMSAT can benefit from larger networks sizes by better modeling the local invariance. We also illustrate that weight-decay does not benefit much from the increased flexibility of neural networks.

For the experiments, we generated a spiral-shaped dataset, each arc containing 300 data points. 
For IMSAT, we used VAT regularization and set $\epsilon = 0.3$ for all the data points.
We compared IMSAT with Deep RIM, which also uses neural networks but with weight-decay regularization. We set the decay rate to 0.0005.
We varied three settings for the network dimensionality of the hidden layers: 5-5, 10-10, and 20-20.

\begin{figure*}[t]
\begin{center}
\centerline{\includegraphics[width=15cm, clip]{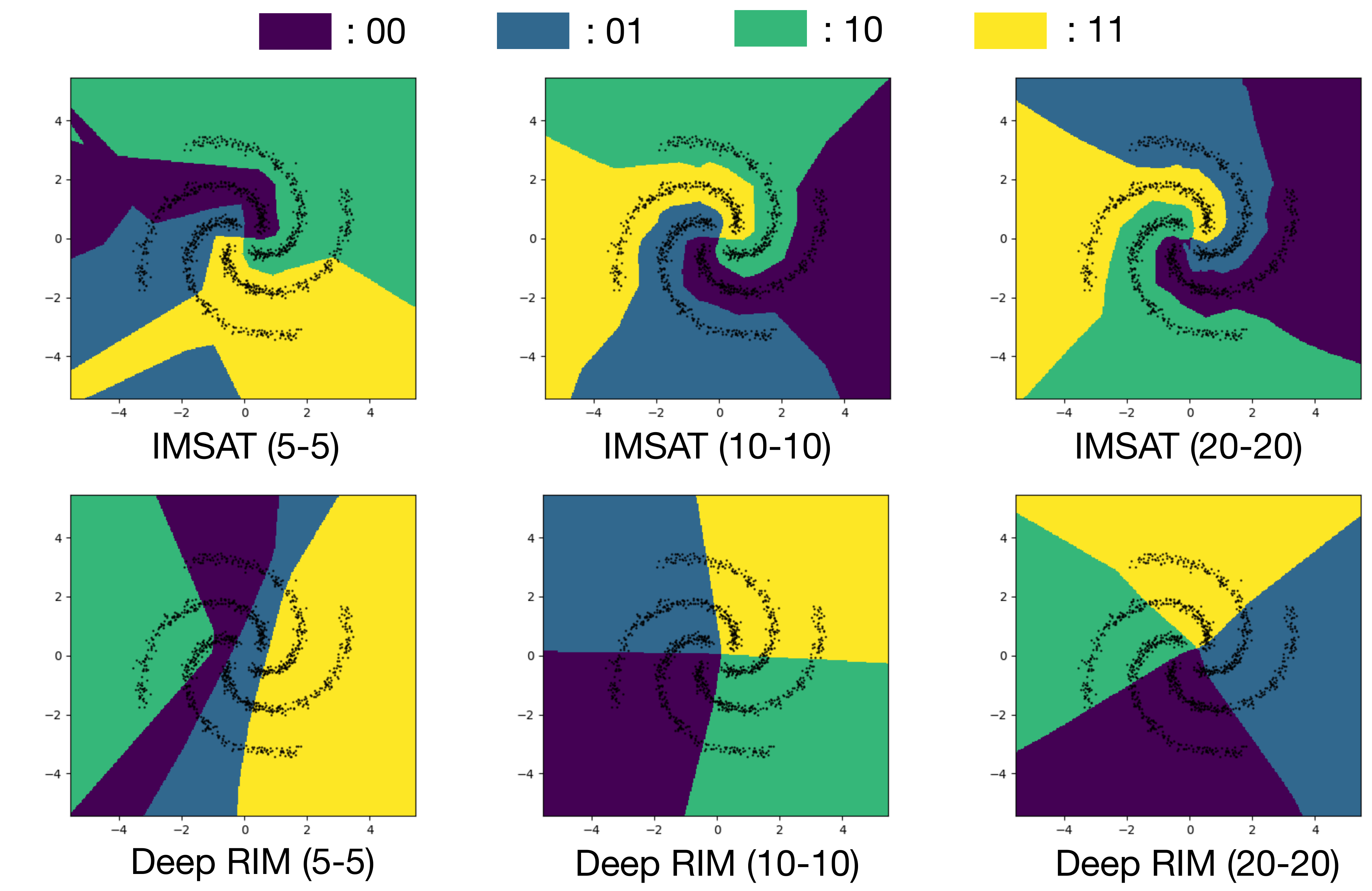}}
\vspace{-0.3cm}
\caption{Comparisons of hash learning with the different regularizations and network sizes using toy datasets.}\label{fig:drim_vat}
\end{center}
\vspace{-0.8cm}
\end{figure*}

Figure \ref{fig:drim_vat} shows the experimental results.
We see that IMSAT (VAT) was able to model the complicated decision boundaries by using the increased network dimensionality. 
On the contrary, the decision boundaries of Deep RIM did not adapt to the non-linearity of data even when the network dimensionality was increased.
This observation may suggest why IMSAT (VAT) benefited from the large networks in the benchmark datasets, while Deep RIM did not.

\end{document}